# Automated detection and counting of windows using UAV imagery based remote sensing

Dhruv Patel[1], Shivani Chepuri[1], Sarvesh Thakur[1], K Harikumar[1], Ravi Kiran S[2], K Madhava Krishna[1]

*Abstract*—Despite the technological advancements in the construction and surveying sector, the inspection of salient features like windows in an under-construction or existing building is predominantly a manual process. Moreover, the number of windows present in a building is directly related to the magnitude of deformation it suffers under earthquakes. In this research, a method to accurately detect and count the number of windows of a building by deploying an Unmanned Aerial Vehicle (UAV) based remote sensing system is proposed. The proposed two-stage method automates the identification and counting of windows by developing computer vision pipelines that utilize data from UAV's onboard camera and other sensors. Quantitative and Qualitative results show the effectiveness of our proposed approach in accurately detecting and counting the windows compared to the existing method.

## I. Introduction and Related Works

UNMANNED AERIAL VEHICLES (UAVS) equipped with different sensors are widely used in various tasks related to the inspection of buildings and other civil structures. The majority of these applications include monitoring the ongoing construction process, surveying the area, classification of completed rooftops, and 3D reconstruction [1]–[7]. Among these, the number and area occupied by windows is a key parameter for risk assessment as it directly affects the structural deformity during natural calamities like earthquakes.

In this paper, we demonstrate methods and algorithms to extract window parameters for under-construction or completely constructed buildings using micro UAVs. Here, we mainly focus on solving the detection and counting problem of windows. For the window detection problem, there have been primarily two approaches - i) Machine Learning-based and ii) Facade parsing-based. The machine learning-based approaches learn features from a set of training images to predict the window as a bounding box or segmented mask. Facade parsing approaches like DeepFacade [12] segment the facades, which can be helpful in solving the detection and counting problem. However, it expects the scene to comprise the entire facade in a single image for segmentation, which is not possible in the case of high-rise buildings. Moreover, the parsing approaches [10]–[12] limit the solution space by using geometrical properties and prior knowledge. Hence, these approaches are not robust to deal with a variety of facades. [8] addresses this problem by first detecting window keypoints and then extracting relationships among keypoints using pixel-wise identity tags. However, this method only tackles window detection and does not solve the counting problem which is important for risk assessment as discussed previously. Hence, we propose a two-stage method for detecting and counting the number of windows/storeys in the facade. The first stage consists of a Deep Learning network - ShufflenetV2 [8], which is trained to detect windows from facade images using heatmaps, and a post-processing module, which further improves the detection accuracy. The second stage proposes a vertical plane mapping algorithm that uses the position and orientation of the UAV to project the detected windows on an imaginary plane and solve the counting problem of windows/storeys. This avoids the duplication of window count arising due to the overlapping frames captured by the camera mounted on the UAV. It also helps tackle the constraint of capturing the entire facade at once, which was there in [12].

Our contribution is three-fold:
- We improve upon the existing Deep Learning-based window detection method by introducing a post-processing module based on Template Matching and Non-maxima suppression.
- We solve the problem of global window counting for a facade, which was not tackled by the previous methods, through a vertical plane mapping algorithm. It utilizes the position and orientation information of the UAV to map the detected windows onto a global vertical plane.
- Our proposed method is able to inspect building properties such as window to facade area ratio, which is critical for structural deformity.

The paper is organized as follows. The collection of aerial imagery data using UAV is described in section II. The algorithms and methods used in this paper are described in section III. The results are presented in section IV followed by the conclusion and future work in section V.

## II. Data Collection

For this research, a dataset of various facades from IIIT Hyderabad campus located in Hyderabad, India is created. The data required to estimate window and storey parameters is collected using DJI Tello, a micro-UAV [1]. DJI Tello has a fixed orientation front-facing camera with onboard sensors such as a barometer, GPS, inertial measurement unit (IMU), etc. The UAV is maneuvered vertically upwards from the ground,

The authors acknowledge the financial support provided by IHUB, IIIT Hyderabad to carry out this research work under the project: IIIT-H/IHub/Project/Mobility/2021-22/M2-003.

[1] Authors are with Robotics Research Centre and [2] author is with Centre for Visual Information Technology, International Institute of Information Technology, Hyderabad-500032, India.

[1] https://www.ryzerobotics.com/tello

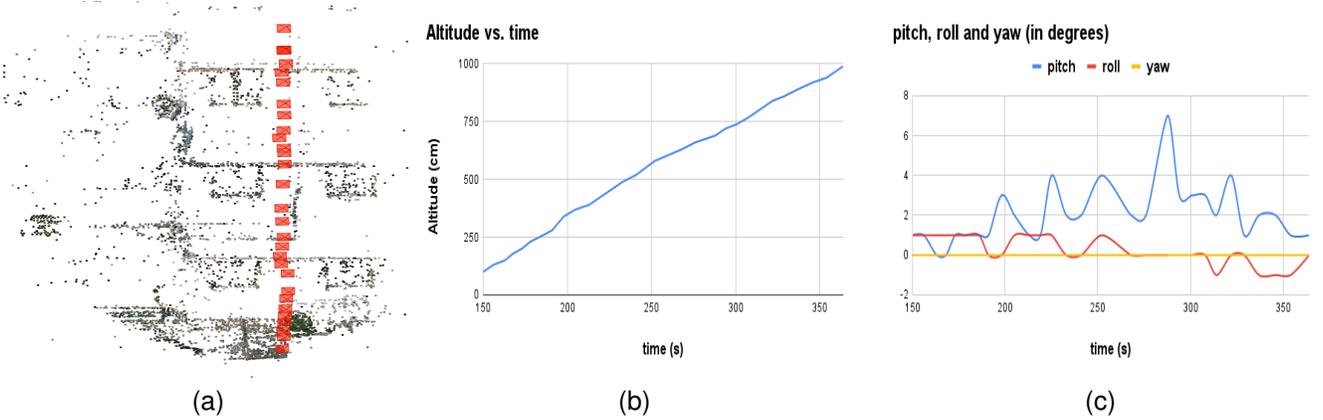

Fig. 1: Bakul building sequence 1. (a) is the Sparse Reconstruction and shows the trajectory of the UAV from the camera poses (in red). (b) shows altitude vs time data of the sequence and (c) shows orientation information with time. Here, note that there is considerable amount of variation in pitch as compared to roll and yaw.

collecting each face of the building. A dataset of around 210 images of 5 buildings: Bakul (B), Vindhya (V), New Faculty Quarters (N), OBH (O), Anand Niwas (A) on the campus has been prepared this way. Fig. 1 shows the trajectory, altitude and orientation information of the UAV for a sample sequence of the data. Fig. 2 shows some collected sample images for this task. Manually annotated windows serve as ground truth for the implemented window detection network.

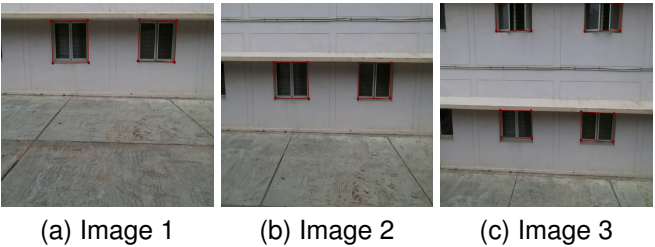

Fig. 2: Data Collection - capturing facade images sequentially for window and storey estimation as the drone moves vertically upwards. Annotated windows are shown in red boxes.

## III. ALGORITHMS AND METHODOLOGY FOR ESTIMATING WINDOW PARAMETERS

This section describes the algorithms and methods implemented to identify the number of windows and stories.

Total window and storey count are crucial structural parameters in building analysis. In this section, we describe the process of estimating the total number of windows and storeys in a facade sequence with high accuracy. We evaluate these parameters in two-stage, window detection, and count estimation. The architecture of the proposed 2-stage method is shown in Fig. 3.

### A. Window detection

*1) Window detection with Deep Learning:* In [8], a heat map-based technique has been presented to detect windows in building facades. The algorithm encodes window key points and their relationships into heat maps which could be learned by an end-to-end network. This helps avoid geometric mismatches providing more robustness in dense window scenarios. We have combined two datasets namely, zju_facade_jcst_-2020 dataset[2] (a collection of about 3000 facade images) and our dataset (a collection of 210 images) of 5 building facades in IIITH campus. This combined dataset [3] is used to train and validate the ShuffleNetV2 backbone network with two fully convolutional heads given in [8].

The training set consists of 3220 images of which 3143 are from *zju* dataset and 77 from the IIIT-H dataset. These 77 images in the training set are from 2 buildings - Bakul and Vindhya. The test set consists of 384 images where 250 are from *zju dataset* and 134 from the IIIT-H dataset (sample shown in Fig. 2). This helps us do rigorous validation and testing of the network on unseen test images with windows of different sizes and shapes. The *zju* dataset consists of images that are captured at larger depths (only facades) when compared to the data captured from UAV. As the dataset appears biased in the color and texture of the facades, it becomes less relevant for UAV surveillance in the Indian context. The pre-trained model (trained only on *zju* dataset) performs poorly on UAV images from the IIIT-H campus. The model performance was improved significantly with a reduction in data bias and a better generalization by training with the images acquired by the UAV as given in Table I. Fig. 4 shows the performance of the model after training with IIIT-H data.

However, a window could still be undetected by the model in the cases of occlusion or distortion (for instance, when a window is near an edge in the image). Intense fine-tuning is not adapted to maintain the generalization of the model. Moreover, the entire window detection stage works frame-wise and does not give any idea on estimation of total window/storey count. This limitation is overcome by the post-processing module in the next section.

[2]link to zju_facade_2020 dataset
[3]link to IIIT-H window dataset

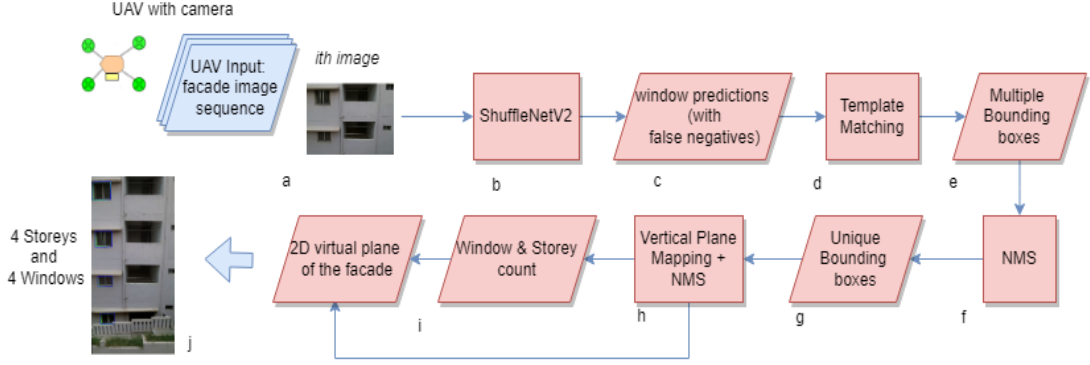

Fig. 3: Block Diagram for window and storey count estimation: Stage (a) represents an image at some index *i* of the input sequence, (b) represents the Deep Learning stage(III-A1), (c) to (g) represents the post-processing module(III-A2), (h) and (i) represent the Count Estimation stage(III-B), (i) represents the final output of the module, (j) shows the entire facade with all windows accurately detected (results shown in 8b)

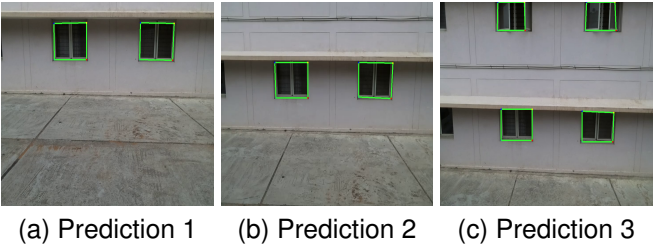

(a) Prediction 1  (b) Prediction 2  (c) Prediction 3

Fig. 4: (a),(b),(c) are the inference from ShuffleNetV2 network when applied on the respective images shown in Fig.2.

*2) Post-processing:* We introduce a post-processing module based on template matching and Non-maximum Suppression (NMS) to improve upon the window detection from model inference. Considering the network detecting at least one window (as the network is reasonably accurate) in a given image, all the detected windows and their variations (rotations of -2.5 and +2.5 degrees) are considered as templates one at a time to identify other windows of the same storey. As a storey usually maintains structural uniformity in its windows, we can optimize this process by limiting search only over the horizontal patch in the same storey.

The variations in rotation as described earlier, account for the windows in the query image whose apparent shape and size are different from that of the template. In this process, different templates can detect the same window many times. Hence, to get the final best detection, we apply NMS to filter out multiple instances of the same window in a given image. NMS selects the prediction with maximum intersection over union (IoU) given in Equation 1 ([9]) and it suppresses all the other bounding boxes based on a threshold. Fig. 7 shows the results of template matching with NMS on images from three different sequences.

$$IoU(B_1, B_2) = \frac{B_1 \cap B_2}{B_1 \cup B_2} \quad (1)$$

where $B_1$ and $B_2$ are any two bounding boxes.

### B. Vertical Plane Mapping and Count Estimation

To visualize all the windows on a common plane parallel to the facade, we propose Algorithm 3 which maps the window coordinates from each image in a building sequence on a common vertical plane. All window placements in the new vertical plane are relative to the first window positioned on the plane. The vertical coordinate ($y$) of the corner pixels in each detected window bounding box is adjusted for the relative pitch angle ($\beta$) of the UAV from the pitch angle of the first image. This information is extracted from the UAV's IMU data, synchronized with the image timestamp. The pitch-corrected vertical offset ($y_c$) is calculated as given in Equation (2).

$$y_c = -D \times \tan\beta \quad (2)$$

where $D$ is the perpendicular distance of the UAV from the facade (Fig. 5) and is a known constant as the UAV captures the entire vertical sequence from a constant depth D. In STEP 2, the Optical Center offset is calculated by shifting the origin to centre of the image i.e. h/2 height (subtracting y coordinate from h/2) and adding pitch correction to it. $y_s$ is the shifted y-coordinate after correcting for pitch and optical center. In STEP 4, using similar triangles and accounting for the UAV height $H$, $Y$ (distance of the window from the line of intersection of the facade plane and the ground plane in the vertical direction) is computed.

In the current approach, there is an assumption that the optical center of the UAV's camera is at the center of the image plane. We consider for pitch angle offset ($\beta$) of the plane as described in STEP 2, as the IMU data from the UAV seem to be significant in pitch angle ($\beta \approx 0.12$ rad = 6.88 degrees) whereas the roll angle ($\gamma \approx 0$ rad) and yaw angle ($\alpha \approx 0$ rad) are negligible. Regardless, this approach can be extended to general conditions without the above assumptions if roll angle ($\gamma$) and yaw angle ($\alpha$) are considerable, unlike our case.

In Fig. 6, it can be seen that as multiple instances of a single window are registered from different images in the sequence, the proposed 2D mapping results in multiple overlapped bounding boxes. NMS approach, as discussed earlier, is utilized again to filter out these boxes and obtain a unique non-overlapping bounding box per window. Fig. 6b displays the final result from the Vertical Plane mapping algorithm. From this approach, the building's window structure, window count, and storey count can be correctly estimated. For storey counting, a sweeping horizontal line starting from the bottom towards the top increments stories count based on the number of times it comes across unique windows horizontally. The entire pipeline for window detection, window counting and storey counting is shown in Fig. 3.

**Algorithm 1** Vertical Plane Mapping Algorithm

**Input:**
(Refer Fig. 5 for details)
$(x,y)$: Window corner coordinates in the image
$H$: Height of the UAV ($m$)
$D$: Depth ($m$)
$f$: Focal Length (pixels)
$h$: Image Height (pixels)
$\beta$: pitch of the UAV relative to the first image in the sequence (radians)

**Output:** (X,Y): Mapped coordinate of the window corner on 2D vertical plane

1 STEP 1: $y_c \leftarrow -D*tan(\beta)$ // Pitch correction
2 STEP 2: $y_s \leftarrow h/2 - y + y_c$ // Optical center correction
3 STEPS 3,4: $X \leftarrow x$, $Y \leftarrow (y_s*D/f)+H$ // Final height

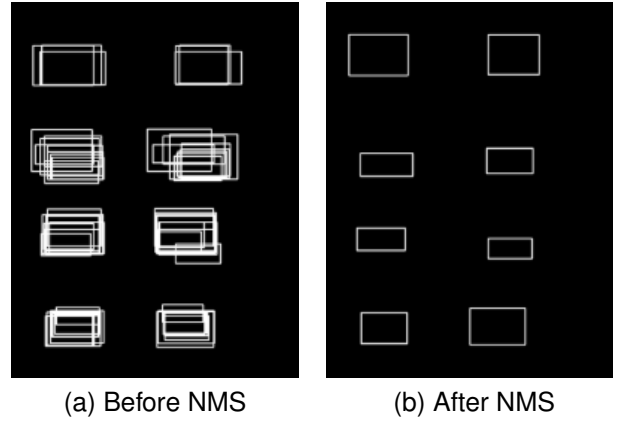

(a) Before NMS  (b) After NMS

Fig. 6: Results of Vertical Plane Mapping Algorithm. NMS here is from stage (h) in Fig. 3. The white rectangles are the bounding boxes of windows from the entire sequence with X,Y value estimated from Algorithm 1.

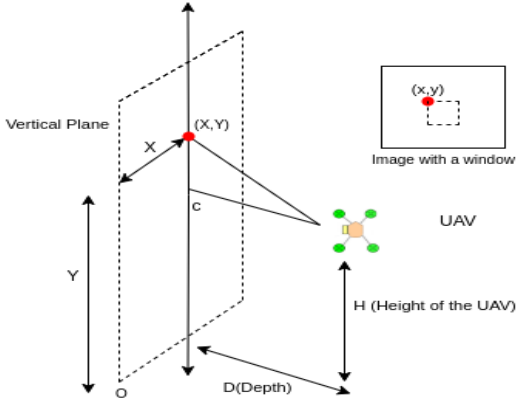

Fig. 5: Mapping the window coordinates of an image to a vertical plane. Coordinates $(x,y)$ refer to the pixel location of a window corner in the image plane. $(X,Y)$ is the coordinate of $(x,y)$ on the vertical Plane. $H$ is the UAV's altitude whereas $D$ is the perpendicular distance of the UAV from the facade.

## IV. EXPERIMENTAL RESULTS AND DISCUSSION

The image data used in this section is captured using the DJI Tello UAV as explained in section II for five multi-storeyed buildings. The results presented here are for (A) Improving the window detection performance, (B) Counting Windows/Storeys, and estimating the ratio of window area to the facade area.

### A. Improvement in Window detection

The improvement in precision and recall while training the model inference module with the collected data set when compared to the pre-trained model given in the literature ([8]) are shown in Table I. Training Shufflenetv2 on the combined zju_facade_jcst2020 and IIIT-H dataset helps in improving the generalization on the test set. From Fig. 7, it can be observed that the post-processing module is able to detect the missed windows (by model) in all three building image sequences.

| Building | AP | Precision | Recall | AP | Precision | Recall |
|---|---|---|---|---|---|---|
| | | Pre-trained | | | After training | |
| N | 0.851 | 0.989 | 0.852 | 0.860 | 0.961 | 0.864 |
| A | 0.823 | 0.908 | 0.824 | 0.834 | 0.968 | 0.835 |
| O | 0.826 | 0.975 | 0.827 | 0.886 | 0.994 | 0.886 |
| B | 0.644 | 0.898 | 0.646 | 0.937 | 0.963 | 0.939 |
| V | 0.737 | 0.933 | 0.737 | 0.995 | 0.905 | 1.000 |
| All | 0.828 | 0.967 | 0.830 | 0.860 | 0.964 | 0.864 |

TABLE I: ShuffleNetV2 performance metrics for window detection on the IIIT-H test set. AP is Area under PR (Precision-Recall) curve. NOTE: Details of the test set are mentioned in III-A1

| Building | Accuracy (Model Inference) | Accuracy (Post-processing) |
|---|---|---|
| B | 98.91% | 99.45% |
| V | 92% | 96% |
| N | 95.76% | 97.46% |
| O | 90.59% | 100% |
| A | 89% | 90% |

TABLE II: Window detection percentage Accuracy (Model inference and after post-processing) for following buildings: Bakul(B), Vindhya(V), New Faculty Quarters(N), OBH(O) and Anand Niwas(A).

From Table II, it is evident that the window detection on the test set improves with post-processing module. Currently, it detects for missed windows through template matching, and hence is limited to buildings with same facade texture throughout the building. In future works, we plan to address it to work on buildings with varying facade texture (eg. Anand Niwas in fig. 7(e), (f)).

### B. Windows, Storeys count and structural properties estimation

Only a single window detection in a horizontal patch is required to correctly estimate the windows count. All detected

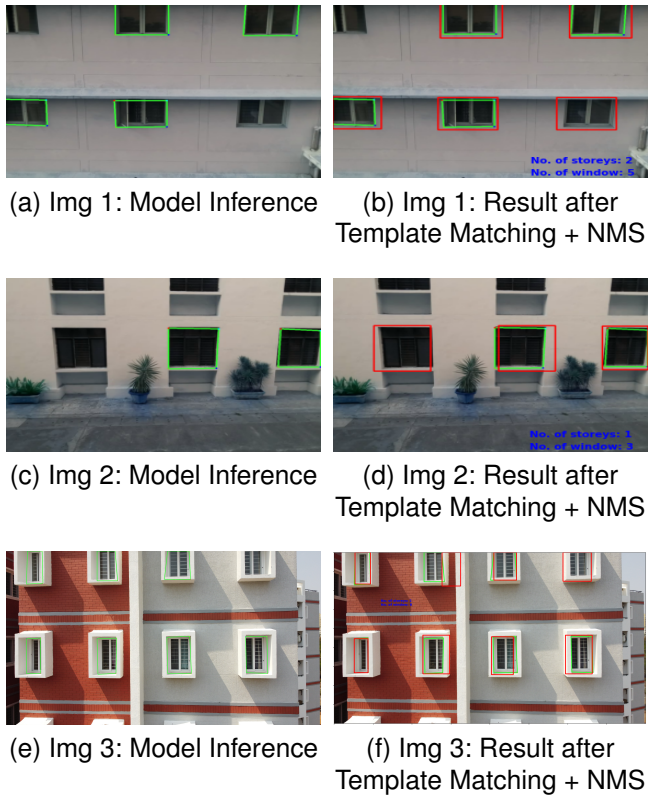

Fig. 7: Post-processing results on ShuffleNetV2 inferred images of Bakul Sequence (a), Vindhya Sequence (c), New Faculty Quarters (e). In (b), (d), and (f), all annotated windows are detected (green). Red boxes represent the final bounding boxes.

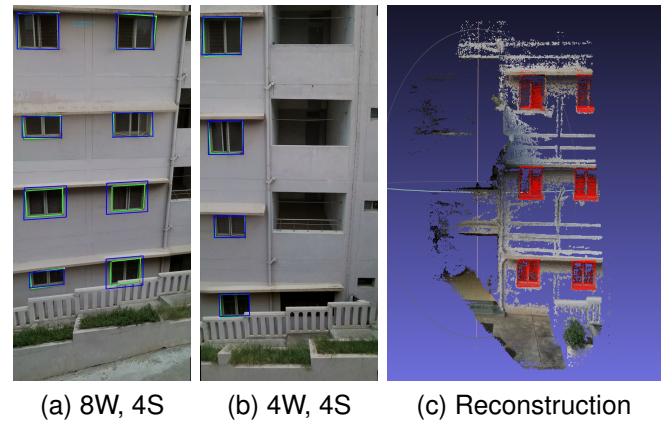

Fig. 8: (a),(b) Panoramic View of facades with Windows (W) and Storeys (S) counts showing no false positives even with the presence of some rectangular structures. Green boxes are predictions from ShuffleNetV2 and blue boxes from Vertical Plane Mapping Algorithm. (c) 3D reconstruction(COLMAP) with windows segmented

windows are mapped onto a vertical plane as shown in Fig. 6b. As described in section III-B, the total number of unique boxes represents the window count and the total number of horizontal levels of the boxes represents the storey count. A panoramic view of the mapped windows of the Bakul building is shown in Fig. 8. For estimating the percentage of window area w.r.t. facade area, the 3D reconstruction of the facade faces was performed using COLMAP and the window area was segmented out from the overall facade area. For Bakul building, the ratio is 0.167 (Fig. 8c) , i.e. the area occupied by windows is 16.7% of the total facade area. Similarly, such properties can be also evaluated for other buildings.

## V. Conclusion

In this paper, we propose a computer vision pipeline to detect the number of windows, their placement and contribution to the structural deformity for particular importance in disaster management. We also evaluate other salient properties such as the height of the building, number of storeys. In addition, modules don't rely on prior information about the building beforehand which makes it robust for its deployment. Currently, image acquisition happens via UAV sensors but processing is performed offline. For future work, we want to perform this task online using onboard computational power through an additional microprocessor.